# A Comparative Analysis of classification data mining techniques : Deriving key factors useful for predicting students' performance

Muhammed Salman Shamsi*, M. Jhansi Lakshmi

* Correspondence:
Affiliation 1:
salman.shamsi@aiktc.org
Assistant Professor,
Department of Computer Engineering,
Anjuman-I-Islam Kalsekar Technical
Campus (AIKTC), Mumbai University,
Navi Mumbai, India
Affiliation 2:
Scholar, Department of CSE
GIET, JNTU-HYD
Hyderabad, India

**Abstract**

Students opting for Engineering as their discipline is increasing rapidly. But due to various factors and inappropriate primary education in India, failure rates are high. Students are unable to excel in core engineering because of complex and mathematical subjects. Hence, they fail in such subjects. With the help of data mining techniques, we can predict the performance of students in terms of grades and failure in subjects. This paper performs a comparative analysis of various classification techniques, such as Naïve Bayes, LibSVM, J48, Random Forest, and JRip and tries to choose best among these. Based on the results obtained, we found that Naïve Bayes is the most accurate method in terms of students' failure prediction and JRip is most accurate in terms of students' grade prediction. We also found that JRip marginally differs from Naïve Bayes in terms of accuracy for students' failure prediction and gives us a set of rules from which we derive the key factors influencing students' performance. Finally, we suggest various ways to mitigate these factors. This study is limited to Indian Education system scenarios. However, the factors found can be helpful in other scenarios as well.

**Keywords:** data mining, classification, failure, prediction, education

**Introduction**

Education is the key to the prosperity of any nation. India is one of the fastest growing nations in the world with the largest youth of population (Unfpa, 2014). Hence, in order to build a skilled workforce education becomes necessary. Students are opting for the fields such as engineering, science, and technology. Unfortunately due to lack of quality education at primary level (The economist, 2008), socio-economic, psychological and other diverse factors, students' failure rates are high and performance is low. Hence to improve the quality of engineering graduates, such cases of failure and poor performance must be monitored proactively. Data mining provides us with tools to analyze a large set of data to derive meaningful data known as knowledge. This helps us to get an insight of data and to reach a meaningful conclusion. Initially, the applications of data mining were restricted to the business domain but now it is extended to education and is known as Educational Data Mining (EDM). EDM deals with the application of data mining tools and techniques to inspect the data at educational institutions for deriving knowledge (Al-razgan, Al-khalifa & Al-khalifa, 2013).

There are various data mining methods such as classification, clustering, and association to analyze data. Classification is supervised learning method that builds a model to classify a data item into a particular class label. The aim of classification is to predict the future outcome based on the currently available data. In clustering, the data objects are combined into sets of objects known as groups or clusters (Han, Kamber & Pei, c2011; Witten & Frank, c2005). The objects within a cluster or a group are highly similar to each other but are dissimilar to the objects in other clusters (Han, Kamber & Pei, c2011). Dissimilarities and similarities measures are based on the attribute values which describe the objects and often involve distance metrics (Han, Kamber & Pei, c2011). Association rule learning involves finding interesting relations between variables in large databases (Piatetsky-shapiro & Frawley, 1991). The aim of association rule learning is to find strong rules in databases based on the various measures of interestingness (Piatetsky-shapiro & Frawley, 1991). Among these techniques, we are going to use classification techniques. In this study, we would analyze various classification methods such as Naïve Bayes, J48, LibSVM, Random Forest and JRip for their accuracy on the given set of data. Weka tool is used for analysis of data and to build the classification model. Weka provides us with a set of machine learning algorithms which can be used for different data mining tasks. It is an open source tool under GNU license (Waikatoacnz, 2016). Weka provides tools for classification, data pre-processing, regression, association rules mining, clustering, and visualization (Waikatoacnz, 2016).

The goal of this study is as follows
- To obtain the most influencing factors that affect students' performance.
- To find best classification method for students' performance prediction in terms of grade and failure in a course.

**Related work**

Since EDM is one of the popular research fields; there are numerous papers we have gone through for literature review. In this section, we discuss some of the works we found most useful for our study.

Abu tair & El-halees, (2012) in their case study discussed various EDM techniques to improve students' performance. Data was collected from the college of Science and Technology – Khanyounis for 15years [1993 to 2007]. This data set consists of 3360 records and 18 attributes. They have used various techniques such as association rule mining, classification, outlier detection and clustering to identify the various factors that affect students' performance. In classification techniques, they have used Rule induction and Naïve Bayes methods. By applying rule-based induction they are able to obtain 71.25% of accuracy whereas by using Naïve Bayes it was reduced to 67.50%. According to rules obtained in the study, factors such as Secondary_School_Type, Matriculation_GPA, City, Gender, and Specialty are factors that affect the grade of the students.

Agarwal, Pandey & Tiwari, (2012) suggested that student's placement is based on his performance in qualifying examination and test marks. They have used attributes such as MAT score, verbal ability score and quantitative ability score to build a decision tree of the student data for placement prediction. LibSVM algorithm with Radial Basis Kernel is used to achieve the overall accuracy of around 97.3%. Since placement is one of the most important parameters for quality of education, it is immensely necessary that students' performance must be improved which is our area of focus throughout the paper.

Kamal, Chowdhury & Nimmy, (2012) had used enrollment data to predict the dropout of Information Systems students studying in the department of Computer Science and Engineering (CSE), University of Chittagong. They had collected data of 1200 students. They had used Bayes theorem based on the knowledge base to predict the dropout. According to the Bayes rules obtained in the study, the most important factors for students' dropout/failure are financial conditions, age group, and gender. They concluded that even though the demographic information is significant to the outcome of the study but the background information such as disability gathered during the enrollment of students is not sufficient for failure prediction.

Márquez-vera, Morales & Soto, (2013) demonstrated that the most influential factors for students' failure are Poor or Not Present in Physics and Math; Not Present in Humanities and Reading & Writing; Poor in English and Social Science; students with more than 15 years age and regular level of motivation. Classification results were obtained for four cases (1) By using all attributes (2) By using best attributes (3) By using Data Balancing (4) By using Cost-Sensitive classification. Among all of the methods used for the study, ADTree was one of the top performers while others were Prism, JRip, and OneR. Some of the limitations of this study are that it puts less focus on socio-economic aspects and it is conducted on school children's hence some of the important factors with respect to higher education had been missed. However, it does convey significant information which can be used to improve the performance at primary level, which in turn will surely improve the performance at a higher level.

Al-barrak & Al-razgan, (2016) suggested using a J48 algorithm to predict final GPA of the students. This study used data of 236 female students who graduated in the year 2012 from Computer Sciences College at King Saud University. This paper attempted to find which courses of previous semesters have a direct impact on final GPA. In result, it was found out that Java1, Database Principles, Software Engineering I, Information Security, Computer Ethics, and Project 2 are most important courses affecting the final GPA of the students.

Bhardwaj & Pal, (2011) in their study gathered data from different institutes affiliated to Dr. R. M. L. Awadh University, Faizabad, India. The sample size was 300 and restricted to Bachelor of Computer Applications (BCA) course.

It was found out using Bayesian classification that student SSC (metric) grade, living location, medium of instruction, mother's qualification, student's habits and type of family are the most important factors for the student's performance.

Kabakchieva, (2013) demonstrated that J48 performance is the best, followed by JRip and k-NN classifiers. The Bayes classifiers are found to be less accurate than the rest. It was found that university admission score and number of failures (in courses) at first-year university exams are the factors that affect students' performance. However, all the tested classifiers had overall accuracy less than 70 % which implies that the error rate was high and the predictions are unreliable.

Mythili & Shanavas, (2014) concludes that attendance, parent's education, locality, gender, economic status are the high potential parameters affecting students' performance in the examination. They have analyzed classifiers such as J48, Random Forest, Multilayer Perceptron, IB1 and Decision Table. It is was found that random forest was the most accurate classifier with an overall accuracy of 89.23%. It also takes less time to build the model than any other classifier.

Al-radaideh, Al-shawakfa & Al-najjar, (2006) in their paper attempted to find out the main parameters which affect students' performance in a particular course. They have used CRISP framework for data mining for this purpose. ID3, C4.5 decision tree, and Naive Bayes were compared. C4.5 was found to be better than others with a classification accuracy of around 38.0531% using 10-fold cross validation. As evident from preceding sentence, it was found that the classification accuracy of the top three algorithms was not so high, hence they concluded that collected sample and attributes are not sufficient to generate a classification model of high quality.

Pandey & Sharma, (2013) compared classification methods to predict grades for undergraduate engineering students. Data was collected from Manav Rachna College of Engineering, Faridabad Dist, Haryana state, India and comprise of 524 instances. In their experiments, they have used J48, Simple Cart, Reptree and NB tree classifiers. J48 was found out to be the most accurate with an overall accuracy of 82.58% using percentage split method and 80.15% using cross-validation.

Ramaswami & Bhaskaran, (2010) developed a CHAID prediction model to predict the performance of students at higher secondary school level. 35 parameters were used for the study. 1000 dataset were collected from five schools in three districts of Tamil Nadu state, India, out of these 772 instances were used after pre-processing of data. The overall accuracy of the model was 44.69% which is better than the model obtained by Al-radaideh, Al-shawakfa & Al-najjar, (2006), but still not acceptable for quality prediction. As per their study, it was found that marks obtained in secondary school level, school location, medium of instruction and type of living are the most influencing factors that affect students' performance.

Kovacic, (2010) used demographic data (gender, age, ethnicity, disability etc.) and study environment data (course program & course block / semester) to predict whether a student will pass or fail in the given course. Classification Trees methods such as CHAID and CART were used with the overall accuracy of 59.4% and 60.5% respectively. According to the trees obtained in the result, it was concluded that ethnicity, course program and course block are the most important factors that help to distinguish between successful and unsuccessful students. As the study is based on enrollment data it leaves important parameters such as assignment marks, test marks, previous academic scores etc which can highly affect students performance.

As evident from above works, we can say that most of the models obtained are of low accuracy (less than 70 %). Also, most of the study has either used enrollment data, demographic information, academics marks or a combination of such data. No study above tends to analyze the effect of psychological, socio-economic or other diverse factors. To overcome these limitations, we have used diverse parameters which include academic parameters, psychological factors, parameters related to socio-economic conditions, background information, demographics data and other diverse factors of students. Hence in this study, we also try to find out whether these all factors together affect students' performance and how.

## Methodology

### Data Collection

In this study, we have taken the data of students belonging to Anjuman-I-Islam's Kalsekar Technical Campus, Navi Mumbai which is affiliated to Mumbai University. The parameters evaluate academics, socio-economic state, psychological state etc. Parameters taken for this study are shown in Table 1.

**Table 1** Parameters related to students

| SR NO. | PARAMETERS | DESCRIPTION |
|---|---|---|
| 1 | Rollno | Unique ID of student |
| 2 | course_id | Unique ID of course |
| 3 | Year | Year in which course is taken |
| 4 | Kt | kt is an acronym for Keep Term. It denotes whether the student is failed in the course or not |
| 5 | Att | Denotes the attendance in course |
| 6 | study_hrs | No. of hours a student spent self-studying at home |
| 7 | Health | Health status of the student |
| 8 | Tuition | Denotes whether the student takes private tuition / coaching or not |
| 9 | source_fees | Denotes the source of paying fees |
| 10 | drop_yr | Denotes whether the student had got academic year drop or not |
| 11 | campus_feedback | Denotes the feedback given by the student to overall campus |
| 12 | travel_time | Travelling time in hours |
| 13 | family_type | Family category based on size |
| 14 | annual_income | Yearly income of the family |
| 15 | father_edu | Father's education status |
| 16 | mother_edu | Mother's education status |
| 17 | father_ocup | Father's occupation |
| 18 | mother_ocup | Mother's occupation |
| 19 | Challenges | Issues in the family |
| 20 | Cast | Caste of the student |
| 21 | mother_tongue | Native Language |
| 22 | Orphan | Whether orphan or not |
| 23 | Kts | Current Total Number of keep terms. In other words total number of subjects which are currently backlog to the student as a result of previous failure(s). |
| 24 | Ssc | Elementary / Secondary school level grade |
| 25 | Hsc | Higher Secondary grade |
| 26 | Medium | Language of teaching in the previous institute |
| 27 | Quota | Determine educational quota type |
| 28 | City | Staying location of student |
| 29 | loc_type | Type of location (urban/rural) |
| 30 | Gender | Sex of student |
| 31 | grade_individual | Course grade achieved |
| 32 | Host | Hosteller / Day scholar |
| 33 | Tw | Term work marks |
| 34 | Orpr | Oral / Practical marks |
| 34 | Test | Average Test Marks |

**Data Pre-Processing**

Data is loaded into weka and irrelevant attributes are removed. Attribute filter NumericalToNominal and StringToNominal is applied to convert all the numerical and string attributes to nominal. The chi-squared filter is used for attribute selection with 20-folds cross-validation. We have chosen two sets of the attributes. Set-I for predicting course failure i.e. kt and Set-II for predicting GPA i.e. grade_individual.

Set-I has 28 attributes from above set in Table 1. Attributes such as grade_individual, tw and orpr were removed as they are not relevant to the prediction of dropout because a 0 (zero) grade point in any of these indicates failure, which is off-course undefined before the conduction of actual exam. Attributes such as host, loc_type, quota, and orphan are removed due to low information gain (average merit < 1.0). Set-II has 31 attributes from the set in Table 1. Attributes such as host, loc_type, gender, and orphan were removed due to low information gain (average merit < 1.0). Now this data set is used as our training data set.

**Evaluating the Models**

Classification algorithms such as Naïve Bayes, JRip, LibSVM, J48 and Random Forest are used on training data to build the model which is cross-validated by 10 folds. Table 2 shows detailed accuracy by class and Table 3 shows the confusion matrices for each algorithm. These tables represent the analysis done on the Set-I attributes which are used to classify students based on the kt attribute. Table 2 represent report generated by weka for each algorithm for this set.

As we compare the weighted average [W.Avg] of each algorithm as shown in Table 2 we found that the True Positive Rate (TP) which denotes the overall accuracy of the classifier is highest for Random Forest (0.833) followed by Naïve Bayes (0.828) and JRip (0.800). However, among these three methods, the False Positive (FP) Rate is highest in JRip (0.398) followed by Random Forest (0.395) and Naïve Bayes (0.268). However, if we look at the CLASS (YES) row which is one of the most important information for the study, it is found that in terms of this class value the most accurate classifier is Naïve Bayes (0.687) followed by JRip (0.508) and Random Forest (0.497).

**Table 2** Detailed accuracy by class for Set-I

| Algorithm | TP | FP | Precision | Recall | F-Measure | ROC | Class |
|---|---|---|---|---|---|---|---|
| Naïve Bayes | 0.873 | 0.313 | 0.897 | 0.873 | 0.885 | 0.885 | NO |
|  | 0.687 | 0.127 | 0.634 | 0.687 | 0.660 | 0.885 | YES |
| [W.avg] | 0.828 | 0.268 | 0.833 | 0.828 | 0.830 | 0.885 |  |
| LibSVM | 1.000 | 1.000 | 0.757 | 1.000 | 0.862 | 0.500 | NO |
|  | 0.000 | 0.000 | 0.000 | 0.000 | 0.000 | 0.500 | YES |
| [W. Avg] | 0.757 | 0.757 | 0.574 | 0.757 | 0.653 | 0.500 |  |
| JRip | 0.894 | 0.492 | 0.850 | 0.894 | 0.871 | 0.713 | NO |
|  | 0.508 | 0.106 | 0.605 | 0.508 | 0.552 | 0.713 | YES |
| [W. Avg] | 0.800 | 0.398 | 0.791 | 0.800 | 0.794 | 0.713 |  |
| J48 | 0.939 | 0.749 | 0.797 | 0.939 | 0.862 | 0.691 | NO |
|  | 0.251 | 0.061 | 0.570 | 0.251 | 0.349 | 0.691 | YES |
| [W. Avg] | 0.772 | 0.582 | 0.742 | 0.772 | 0.738 | 0.691 |  |
| Random Forest | 0.941 | 0.503 | 0.854 | 0.941 | 0.895 | 0.888 | NO |
|  | 0.497 | 0.059 | 0.730 | 0.497 | 0.591 | 0.888 | YES |
| [W. Avg] | 0.833 | 0.395 | 0.824 | 0.833 | 0.822 | 0.888 |  |

Table 3 shows the confusion matrices which denotes the number of tuples classified in class a [NO], b [YES] and c [NA] out of total instances of 1476.

**Table 3** Confusion matrices for Set-I

| Algorithm | Actual Class | Classified Class ||
|---|---|---|---|
|  |  | A [NO] | B [YES] |
| Naïve Bayes | A [NO] | 976 | 142 |
|  | B [YES] | 112 | 246 |
| LibSVM | A [NO] | 1,118 | 0 |
|  | B [YES] | 358 | 0 |
| JRip | A [NO] | 999 | 119 |
|  | B [YES] | 176 | 182 |
| J48 | *A [NO]* | 1,050 | 68 |
| | *B [YES]* | 268 | 90 |
| Random forest | *A [NO]* | 1,052 | 66 |
| | *B [YES]* | 180 | 178 |

Table 4 and Table 5 represent the analysis done on the Set-II attributes using weka tool which is used to obtain model(s) to predict the theory grade of students represented by the grade_individual attribute. This attribute represents the grading point of the subject whose values can vary from 0 to 10 which indicates the corresponding grade obtained in the subject. Note that the value of -1 indicates that the course does not have theory exam. Hence the theory exam grade is not applicable.

As we compare the weighted average [W.Avg] of each algorithm as shown in Table 4, we found that the True Positive Rate (TP) is highest for JRip (0.594) followed by J48 (0.586) and Random Forest (0.554).

**Table 4** Detailed accuracy by class for Set-II

| Algorithm | TP | FP | Precision | Recall | F-Measure | ROC | Class |
|---|---|---|---|---|---|---|---|
| Naïve Bayes | 0.887 | 0.015 | 0.829 | 0.887 | 0.857 | 0.988 | -1 |
| | 0.893 | 0.046 | 0.849 | 0.893 | 0.870 | 0.977 | 0 |
| | 0.522 | 0.187 | 0.557 | 0.522 | 0.539 | 0.778 | 4 |
| | 0.138 | 0.085 | 0.172 | 0.138 | 0.153 | 0.701 | 5 |
| | 0.286 | 0.123 | 0.282 | 0.286 | 0.284 | 0.721 | 6 |
| | 0.361 | 0.096 | 0.249 | 0.361 | 0.295 | 0.784 | 7 |
| | 0.114 | 0.010 | 0.211 | 0.114 | 0.148 | 0.709 | 8 |
| | 0.000 | 0.011 | 0.000 | 0.000 | 0.000 | 0.675 | 9 |
| | 0.200 | 0.008 | 0.214 | 0.200 | 0.207 | 0.738 | 10 |
| [W.Avg] | 0.524 | 0.105 | 0.518 | 0.524 | 0.519 | 0.820 | |
| LibSVM | 0.148 | 0.000 | 1.000 | 0.148 | 0.258 | 0.574 | -1 |
| | 0.961 | 0.029 | 0.907 | 0.961 | 0.933 | 0.966 | 0 |
| | 0.967 | 0.649 | 0.401 | 0.967 | 0.567 | 0.659 | 4 |
| | 0.000 | 0.000 | 0.000 | 0.000 | 0.000 | 0.500 | 5 |
| | 0.000 | 0.000 | 0.000 | 0.000 | 0.000 | 0.500 | 6 |
| | 0.000 | 0.000 | 0.000 | 0.000 | 0.000 | 0.500 | 7 |
| | 0.000 | 0.000 | 0.000 | 0.000 | 0.000 | 0.500 | 8 |
| | 0.000 | 0.000 | 0.000 | 0.000 | 0.000 | 0.000 | 9 |
| | 0.000 | 0.000 | 0.000 | 0.000 | 0.000 | 0.500 | 10 |
| [W. Avg] | 0.530 | 0.208 | 0.408 | 0.530 | 0.408 | 0.661 | |
| JRIP | 0.974 | 0.009 | 0.903 | 0.974 | 0.937 | 0.975 | -1 |
| | 0.979 | 0.026 | 0.916 | 0.979 | 0.947 | 0.972 | 0 |
| | 0.945 | 0.533 | 0.444 | 0.945 | 0.604 | 0.705 | 4 |
| | 0.000 | 0.002 | 0.000 | 0.000 | 0.000 | 0.664 | 5 |
| | 0.000 | 0.002 | 0.000 | 0.000 | 0.000 | 0.672 | 6 |
| | 0.025 | 0.001 | 0.750 | 0.025 | 0.049 | 0.660 | 7 |

|  | 0.029 | 0.003 | 0.200 | 0.029 | 0.050 | 0.581 | 8 |
|---|---|---|---|---|---|---|---|
|  | 0.000 | 0.000 | 0.000 | 0.000 | 0.000 | 0.643 | 9 |
|  | 0.000 | 0.002 | 0.000 | 0.000 | 0.000 | 0.472 | 10 |
| [W. Avg] | 0.594 | 0.173 | 0.481 | 0.594 | 0.480 | 0.767 |  |
| J48 | 0.904 | 0.009 | 0.897 | 0.904 | 0.900 | 0.949 | -1 |
|  | 0.949 | 0.018 | 0.938 | 0.949 | 0.944 | 0.974 | 0 |
|  | 0.959 | 0.558 | 0.436 | 0.959 | 0.599 | 0.767 | 4 |
|  | 0.000 | 0.003 | 0.000 | 0.000 | 0.000 | 0.696 | 5 |
|  | 0.000 | 0.002 | 0.000 | 0.000 | 0.000 | 0.693 | 6 |
|  | 0.034 | 0.000 | 1.000 | 0.034 | 0.065 | 0.718 | 7 |
|  | 0.000 | 0.001 | 0.000 | 0.000 | 0.000 | 0.683 | 8 |
|  | 0.000 | 0.001 | 0.000 | 0.000 | 0.000 | 0.651 | 9 |
|  | 0.000 | 0.000 | 0.000 | 0.000 | 0.000 | 0.773 | 10 |
| [W. Avg] | 0.586 | 0.179 | 0.499 | 0.586 | 0.476 | 0.802 |  |
| Random Forest | 0.948 | 0.013 | 0.858 | 0.948 | 0.901 | 0.995 | -1 |
|  | 0.928 | 0.050 | 0.845 | 0.928 | 0.885 | 0.985 | 0 |
|  | 0.716 | 0.312 | 0.508 | 0.716 | 0.594 | 0.762 | 4 |
|  | 0.084 | 0.046 | 0.819 | 0.084 | 0.116 | 0.650 | 5 |
|  | 0.164 | 0.103 | 0.212 | 0.164 | 0.185 | 0.675 | 6 |
|  | 0.160 | 0.037 | 0.275 | 0.160 | 0.202 | 0.768 | 7 |
|  | 0.057 | 0.007 | 0.167 | 0.057 | 0.085 | 0.745 | 8 |
|  | 0.000 | 0.005 | 0.000 | 0.000 | 0.000 | 0.747 | 9 |
|  | 0.000 | 0.005 | 0.000 | 0.000 | 0.000 | 0.864 | 10 |
| [W. Avg] | 0.554 | 0.133 | 0.494 | 0.554 | 0.514 | 0.806 |  |

Table 5 shows the confusion matrices which denotes the number of tuples classified in class -1 to 10 class out of total instances of 1476.

**Table 5** Confusion Matrices for Set-II

| Algorithm | Actual Class | Classified Class | | | | | | | | |
|---|---|---|---|---|---|---|---|---|---|---|
|  |  | *A[-1]* | *B [0]* | *C [4]* | *D [5]* | *E [6]* | *F [7]* | *G [8]* | *H [9]* | *I[10]* |
| Naïve Bayes | *A [-1]* | 102 | 0 | 2 | 1 | 0 | 5 | 2 | 1 | 2 |
|  | *B [0]* | 4 | 299 | 21 | 4 | 2 | 3 | 0 | 2 | 0 |
|  | *C [4]* | 5 | 40 | 239 | 59 | 77 | 30 | 2 | 4 | 2 |
|  | *D [5]* | 0 | 7 | 72 | 23 | 37 | 22 | 0 | 3 | 3 |
|  | *E [6]* | 1 | 4 | 69 | 27 | 61 | 44 | 3 | 1 | 3 |
|  | *F [7]* | 6 | 1 | 23 | 12 | 27 | 43 | 5 | 2 | 0 |
|  | *G [8]* | 1 | 1 | 2 | 4 | 6 | 15 | 4 | 2 | 0 |
|  | *H [9]* | 2 | 0 | 1 | 2 | 5 | 6 | 2 | 0 | 1 |
|  | *I [10]* | 2 | 0 | 0 | 2 | 1 | 5 | 1 | 1 | 3 |
| LibSVM | *A [-1]* | 17 | 4 | 94 | 0 | 0 | 0 | 0 | 0 | 0 |
|  | *B [0]* | 0 | 322 | 13 | 0 | 0 | 0 | 0 | 0 | 0 |
|  | *C [4]* | 0 | 15 | 443 | 0 | 0 | 0 | 0 | 0 | 0 |

|  | D [5] | 0 | 3 | 164 | 0 | 0 | 0 | 0 | 0 | 0 |
|---|---|---|---|---|---|---|---|---|---|---|
|  | E [6] | 0 | 3 | 210 | 0 | 0 | 0 | 0 | 0 | 0 |
|  | F [7] | 0 | 4 | 115 | 0 | 0 | 0 | 0 | 0 | 0 |
|  | G [8] | 0 | 3 | 32 | 0 | 0 | 0 | 0 | 0 | 0 |
|  | H [9] | 0 | 1 | 18 | 0 | 0 | 0 | 0 | 0 | 0 |
|  | I [10] | 0 | 0 | 15 | 0 | 0 | 0 | 0 | 0 | 0 |
| JRip | A [-1] | 112 | 0 | 3 | 0 | 0 | 0 | 0 | 0 | 0 |
|  | B [0] | 0 | 328 | 7 | 0 | 0 | 0 | 0 | 0 | 0 |
|  | C [4] | 2 | 19 | 433 | 2 | 2 | 0 | 0 | 0 | 0 |
|  | D [5] | 0 | 3 | 162 | 0 | 1 | 0 | 0 | 0 | 1 |
|  | E [6] | 0 | 3 | 206 | 0 | 0 | 0 | 1 | 0 | 1 |
|  | F [7] | 4 | 2 | 107 | 1 | 0 | 3 | 1 | 0 | 1 |
|  | G [8] | 0 | 3 | 30 | 0 | 0 | 1 | 1 | 0 | 0 |
|  | H [9] | 2 | 0 | 15 | 0 | 0 | 0 | 2 | 0 | 0 |
|  | I [10] | 2 | 0 | 13 | 0 | 0 | 0 | 0 | 0 | 0 |
| J48 | A [-1] | 104 | 1 | 10 | 0 | 0 | 0 | 0 | 0 | 0 |
|  | B [0] | 2 | 318 | 13 | 1 | 1 | 0 | 0 | 0 | 0 |
|  | C [4] | 2 | 17 | 439 | 0 | 0 | 0 | 0 | 0 | 0 |
|  | D [5] | 0 | 1 | 164 | 0 | 2 | 0 | 0 | 0 | 0 |
|  | E [6] | 2 | 2 | 207 | 2 | 0 | 0 | 0 | 0 | 0 |
|  | F [7] | 3 | 0 | 112 | 0 | 0 | 4 | 0 | 0 | 0 |
|  | G [8] | 0 | 0 | 32 | 1 | 0 | 0 | 0 | 2 | 0 |
|  | H [9] | 1 | 0 | 17 | 0 | 0 | 0 | 1 | 0 | 0 |
|  | I [10] | 2 | 0 | 13 | 0 | 0 | 0 | 0 | 0 | 0 |
| Random Forest | A [-1] | 109 | 0 | 3 | 0 | 2 | 1 | 0 | 0 | 0 |
|  | B [0] | 0 | 311 | 22 | 0 | 1 | 1 | 0 | 0 | 0 |
|  | C [4] | 4 | 39 | 328 | 25 | 51 | 11 | 0 | 0 | 0 |
|  | D [5] | 0 | 9 | 109 | 14 | 25 | 6 | 1 | 1 | 2 |
|  | E [6] | 3 | 6 | 127 | 18 | 35 | 15 | 4 | 2 | 3 |
|  | F [7] | 6 | 2 | 44 | 10 | 34 | 19 | 3 | 0 | 1 |
|  | G [8] | 0 | 1 | 9 | 2 | 8 | 7 | 2 | 5 | 1 |
|  | H [9] | 3 | 0 | 3 | 1 | 6 | 4 | 2 | 0 | 0 |
|  | I [10] | 2 | 0 | 1 | 4 | 3 | 5 | 0 | 0 | 0 |

Table 6 shows the time took to build prediction model by each method on Set-I and Set-II data having total instances of 1476 each. As evident from the table Naïve Bayes takes the shortest amount of time of 0.04 secs and 0 secs to build a model for Set-I and Set-II data respectively.

**Table 6** Time taken for Model Building for Set-I & Set-II

| **Algorithm** | Time Taken to Build Model (in Seconds) | |
|---|---|---|
|  | Set-I | **Set-II** |
| Naïve Bayes | 0.04 | 0 |
| LibSVM | 0.63 | 1.62 |
| JRip | 0.24 | 0.16 |
| J48 | 0.06 | 0.02 |
| Random forest | 0.39 | 0.75 |

## Results

From all the analyses we have done so far, we found out that Random Forest, Naïve Bayes, and JRip are the best three methods with the differences of approximately 0.5 to 2.8 % in accuracy for Set-I data. Among these, we prefer Naïve Bayes because it gives us the most accurate prediction (68.7%) with respect to failure in a course i.e. all the instances which fall in class (YES), which is our aim and it takes the shortest time of 0.04 secs to build a model. Even though Random Forest has highest average accuracy, it lags much behind Naïve Bayes in terms of class (YES) prediction (49.7%), also the difference between their accuracy is only 0.5%. Hence, we prefer Naïve Bayes over Random Forest. For Set-II data, JRip is the most accurate classifier even though it takes slightly more time to build a model than Naïve Bayes and J48. However, the difference of time between JRip and these classifiers (0.12 secs) is quite negligible. This establishes our second objective.

JRip algorithm also gives us a set of rules which help us to derive the most influential factors which affect students' performance. Fig. 1, shows the rules we have derived by using JRip method for Set-I data. According to these rules, it was found that attendance (att), term test grade (test), travelling time (travel_time), academic year drop (drop_yr), parents annual income (annual_income), metric/ssc percentage (ssc), number of hours spent studying (study_hrs), source of fees (source_fees), number of backlog subjects (kts) and father's education (father_edu) are the factors which determine whether a student will fail in the subject or not. Among these, term test grade (test) and number of subject kts (kts) have high-frequency count than other attributes. Hence, these are the parameters which highly affect students' performance.

```
(test = 0) => kt=YES (62.0/5.0)

(test = 5) and (campus_feedback = 4) and (study_hrs = 2) => kt=YES (23.0/5.0)

(test = 5) and (ssc = FIRST CLASS) and (source_fees = PARENTS) => kt=YES (27.0/6.0)

(drop_yr = YES) and (travel_time = UPTO 1 HRS) => kt=YES (22.0/6.0)

(test = 6) and (annual_income = LESS THAN 1 LAKH) and (kts = 4) and (drop_yr = NO) => kt=YES (10.0/1.0)

(test = 6) and (travel_time = MORE THAN 3 HOURS) and (health = EXCELLENT) => kt=YES (6.0/0.0)

(kts = 7) => kt=YES (26.0/6.0)

(kts = 5) and (test = 4) => kt=YES (13.0/2.0)

(kts = 3) and (att = AVERAGE) and (father_edu = 1) => kt=YES (23.0/7.0)

(kts = 5) and (campus_feedback = 4) => kt=YES (34.0/14.0)

(course_id = CSC303) and (kts = 4) => kt=YES (10.0/1.0)

=> kt=NO (1220.0/155.0)
```

**Fig. 1** JRip Rules, Set-I attributes for kt prediction

Fig. 2, shows the rules we have derived by using JRip method for Set-II. According to the rules generated, course_id, test, kt, tw, medium and kts are the factors which determine students' grade. Among these course id, term test marks(test) and number of subject kts (kts) are the most influencing factors based on frequency count which affects students' grade. This establishes our first goal.

```
(course_id = CPC501) and (test = 10) => grade_individual=9 (3.0/0.0)

(test = -1) => grade_individual=-1 (115.0/10.0)

(course_id = CPL502) => grade_individual=-1 (10.0/2.0)

(kts = 0) and (course_id = CPL501) => grade_individual=7 (10.0/4.0)

(kts = 0) and (tw = 7) and (medium = VERNACULAR) => grade_individual=7 (12.0/4.0)

(kt = YES) => grade_individual=0 (344.0/22.0)
```

**Fig. 2** JRip Rules, Set-II attributes for grade_individual prediction

# Discussion

In this study, we have analyzed the best classification method as per accuracy and our need. We also found various factors that affect students' performance. In this section, we will analyze these factors and suggest various ways to deal with them.

As we look into the factors described in result section, it's been noticed that it doesn't include any factor such as family issues, cast, gender, health, family type, etc., which is unusual. Considering above point, we can infer following things. First, the students have not disclosed their personal issues due to privacy concern or other factors. Second, the sample we have taken is from a minority institute. Hence, we fail to find caste factor which is less prevalent in such religious minority institutes. Also, most of the instances in sample exhibit homogenous nature in health and family attributes, so we fail to find any co-relation between them and students' performance.

Factors such as parents' annual income and source of fees help us to identify the financial conditions of students. Poor financial conditions can lead to stress among the students (Andrews & Wilding, 2004; Kumar, Dagli, Mathur, Jain, Prabu & Kulkarni, 2009). Hence, stress can degrade students' performance. In such cases, financial aid can be provided to these students. Such funds can be granted from Students Aid Fund which is maintained by the institute or by means of private or government scholarships. This will also reduce the financial burden on their parents.

Attendance is also one of the factors which affect the performance. So, students should be monitored weekly or monthly for the shortage of attendance in any course and the parents must be informed about the same. Wartman & Savage, (2008) supported parents' involvement in the higher education of their adult children. According to their study, parents' involvement is positive because it helps them to support their child, better understand the higher education scenario and its constraints, helps to improve and understand the relationship between parents, students, and institution etc. So, by informing parents about their child's attendance it is expected that they will somehow influence their children to attain lectures on a regular basis and inform the institute about any genuine problems which they or their child encountered that cannot be tackled by them alone. We also recommend making teaching process more student-centric. Different teaching methodologies must be adopted as per students' need. With respect to preceding statement, we imply that instructor should try out different teaching methods and models to improve his teaching style so that lectures become more thought-provoking and interesting. This will lead to regular attendance of the students, which in turn helps in proper concept building of the subject since upcoming topics are highly related to topics already taught.

Certain factors such as the number of hours spent studying, travel time and father's education cannot be directly arbitrated by the institute. However, we can still deal with them. For example, the number of hours spent studying can be improved if students are encouraged to develop good study habits. Kumar, (2015), demonstrated that students who were not able to develop their skills at secondary or primary level tend to proceed towards higher education without evolving the habit of planning their study time, hence they spent less time self-studying. Instructors should motivate students at all levels of education to develop good study habits and can advise them about various ways to overcome their problems. Some of the techniques which students can follow are to organize their time, relate newly acquired knowledge with analogy and previously acquired knowledge, take proper sleep and perform a daily revision of the topics taught in the classroom. Another factor is the travel time, which leads to wastage of precious time and in some cases can be a stress factor among students. To save time, students must be advised to stay at a hostel or self-managed rented accommodation. Day scholars must be guided on how to optimally utilize their available time. Students who are staying very far can opt for online education which will not only save their time but also give them a quality education. Many reputed universities are now offering online degrees which are not only cheaper compared to their on-campus cost but also has good industrial value compared to other conventional universities. Another factor which is beyond our control is father's education. However, it gives us helpful insight about how father's education influences child's performance. If we look at rules obtained in Fig 1. We can say that the students whose fathers are metric pass (father_edu=1) are more likely to fail in the course combined with other factors.

One of the most important factors revealed is the current number of subject kts (backlog) a student is holding. It was found that the students with more than 3 kts combined with others factors are more likely to get failed in a subject. Since university semester in India mostly comprises of around 6 subjects, the students might not be able to cope up with the burden of an extra pending subject(s). To overcome this problem, a careful academic planning is required by the student.

This can be achieved with the help of students' mentorship program. Each faculty can be assigned a set of such students, which they can guide effectively during the whole semester.

Lastly, the students can also undergo counseling to improve their performance. This will also help to obtain hidden factors which are not covered in this study or are not disclosed by the students due to some reason. These factors can be used in the future studies to improve the prediction model in terms of accuracy.

## Conclusion & Future Work

Let us summarize the study we have done so far. The data we collected was from a particular minority institute, but the study is applicable to other institutions for the following reasons. Data such as caste, religion, parents income, parents educational qualification, etc., is already collected from the student at the time of admission in all kinds of institutes in India as per norms and also the parameters found in our study are highly correlated with that of other studies we have gone through. However, some additional factors can be found which we have missed, as already discussed in para 2 of preceding section. Based on the results obtained, we also found that Naïve Bayes is the most effective method compared to other three methods, for Set-I data, it is most accurate with respect to course backlog prediction (kt=YES) i.e. having a TP rate of 64.7% and has most negligible execution time of 0.04 secs. For Set-II data, JRip is the most effective method because it is most accurate and has an execution time of 0.16 secs.

Finally, Naïve Bayes and JRip can be used by any institute for their analytic needs because the application of such methods is independent of the local context of the study. Naïve Bayes can help institutes to predict students' failure in a course and JRip can give factors which influence students' performance in the form of rules. Such information can immensely benefit the Institutes by allowing them to take remedial actions. Institutes can perform analysis on students' data by using weka tool or by implementing the said techniques in an online management system. However, we recommend building an artificial intelligence based academic management system which can predict failure of students in the enrolled course(s) and present the result in a user-friendly format. An online system will be more beneficial since it provides real-time analytics results in the form of tables, charts or any other suitable format as desired by the institute. Such a system is already being tested at Anjuman-I-Islam's Kalsekar Technical Campus (AIKTC) for feasibility. The knowledge obtained from the specified data mining methods will help the instructor to pay more attention to vulnerable students.

If we go through the results obtained on Set-II data, we were only able to achieve an accuracy of 59.4% (JRip) which is not suitable for accurate prediction. Also, the rules obtained in Fig 2, appears mostly trivial but one of them does convey some significant information such as, according to the fifth rule, we can say the students who don't have any backlog subject (kts), whose term work (tw) is 7 grade point and had studied in vernacular medium previously are found to be scoring 7 grade point in that subject. The model obtained correctly classifies 12 instances and there are 4 incorrect classifications for this rule. It is evident that the support for such rule is very less. Hence, we conclude that the current set of data instances are insufficient to create a model of higher accuracy for Set-II data.

In the future study, we will try to collect large and diverse samples to detect other important factors affecting students' performance which we have missed. It would be interesting to know how caste, gender, psychological factors and socio-economic status affect students' performance which we are unable to detect. A future study can be undertaken to study the impact of deep learning methods and how to incorporate them into students' performance prediction. This will help in the building of more accurate prediction models. We also wish to find out how the current teaching methods in Indian Education system influence students' academic performance. However, this study can be of immense benefit to the welfare of students if the parameters obtained are proactively monitored by the institutes. Strict remedial action as suggested can be taken if the student is showing signs of failure or poor grade in a subject. This will lead to better results and quality education by tapping up the potential problem in an early stage.

**Acknowledgment**

I would sincerely like to thank all the students and staff members of AIKTC, Computer Engineering department who participated in this study and took out the time for filling up the data forms. I would also like to thank, Prof. Tabrez Khan, Head, Department of Computer Engineering, AIKTC, for being extremely supportive in the conduction of this study. I would also like to thank, Dr. Mrs. Ravindra Tiwari, Principal Global Institute of Engineering & Technology for providing us a platform to showcase our talent.